\documentclass[sigconf,screen,prologue,dvipsnames,table]{acmart}

\usepackage{epsfig}
\usepackage{graphicx}
\usepackage{amsmath}
\usepackage{amssymb}
\usepackage{bbm}
\usepackage[mathscr]{euscript}
\usepackage{multirow}
\usepackage{booktabs}
\usepackage{algorithm}
\usepackage{algorithmicx}
\usepackage{algpseudocode}
\algnewcommand{\LineComment}[1]{\State \(\triangleright\) #1}
\usepackage{enumitem}% http://ctan.org/pkg/enumitem
\usepackage{subdefs}
\usepackage{ifthen}

\usepackage{colortbl}
\definecolor{Gray}{gray}{0.85}
\definecolor{SkyBlue}{rgb}{0.88,1,1}
\newcolumntype{a}{>{\columncolor{Gray}}c}

\newcommand{\link}[1]{{\color{blue}\href{#1}{#1}}}

\copyrightyear{2021}
\acmYear{2021}
\setcopyright{acmlicensed}\acmConference[MM '21]{Proceedings of the 29th ACM International Conference on Multimedia}{October 20--24, 2021}{Virtual Event, China}
\acmBooktitle{Proceedings of the 29th ACM International Conference on Multimedia (MM '21), October 20--24, 2021, Virtual Event, China}
\acmPrice{15.00}
\acmDOI{10.1145/3474085.3475522}
\acmISBN{978-1-4503-8651-7/21/10}

\acmSubmissionID{1894}

\begin{document}
\fancyhead{}
\title[BinnedCrowd]{Wisdom of (Binned) Crowds: A Bayesian Stratification Paradigm for Crowd Counting}

\author{Sravya Vardhani Shivapuja}
\email{sravya.vardhani@research.iiit.ac.in}
\orcid{1234-5678-9012}
\affiliation{%
  \institution{CVIT, IIIT Hyderabad}
  \city{Hyderabad 500032}
  \country{INDIA}
}
\author{Mansi Pradeep Khamkar}
\email{mansi.khamkar@students.iiit.ac.in}
\affiliation{%
  \institution{CVIT, IIIT Hyderabad}
  \city{Hyderabad 500032}
  \country{INDIA}
}
\author{Divij Bajaj}
\email{divij.bajaj.ece17@itbhu.ac.in}
\affiliation{%
  \institution{CVIT, IIIT Hyderabad}
  \city{Hyderabad 500032}
  \country{INDIA}
}
\author{Ganesh Ramakrishnan}
\email{ganesh@cse.iitb.ac.in}
\affiliation{%
  \institution{Dept. of CSE, IIT Bombay}
  \city{Mumbai 400076}
  \country{INDIA}}

\author{Ravi Kiran Sarvadevabhatla}
\email{ravi.kiran@iiit.ac.in}
\affiliation{%
 \institution{CVIT, IIIT Hyderabad}
 \city{Hyderabad 500032}
 \country{INDIA}
 }

\renewcommand{\shortauthors}{Sravya et al.}

\begin{abstract}
Datasets for training crowd counting deep networks are typically heavy-tailed in count distribution and exhibit discontinuities across the count range. As a result, the de facto statistical measures (MSE, MAE) exhibit large variance and tend to be unreliable indicators of performance across the count range. To address these concerns in a holistic manner, we revise  processes  at various stages of the standard crowd counting pipeline. To enable principled and balanced minibatch sampling, we propose a novel smoothed Bayesian sample stratification approach. We propose a novel cost function which can be readily incorporated into existing crowd counting deep networks to encourage strata-aware optimization. We analyze the performance of representative crowd counting approaches across standard datasets at per strata level and in aggregate. We analyze the performance of crowd counting approaches across standard datasets and demonstrate that our proposed modifications noticeably reduce error standard deviation. Our contributions represent a nuanced, statistically balanced and fine-grained characterization of performance for crowd counting approaches. Code, pretrained models and interactive visualizations can be viewed at our project page \link{deepcount.iiit.ac.in}.
\end{abstract}




\begin{teaserfigure}
\includegraphics[width=\textwidth]{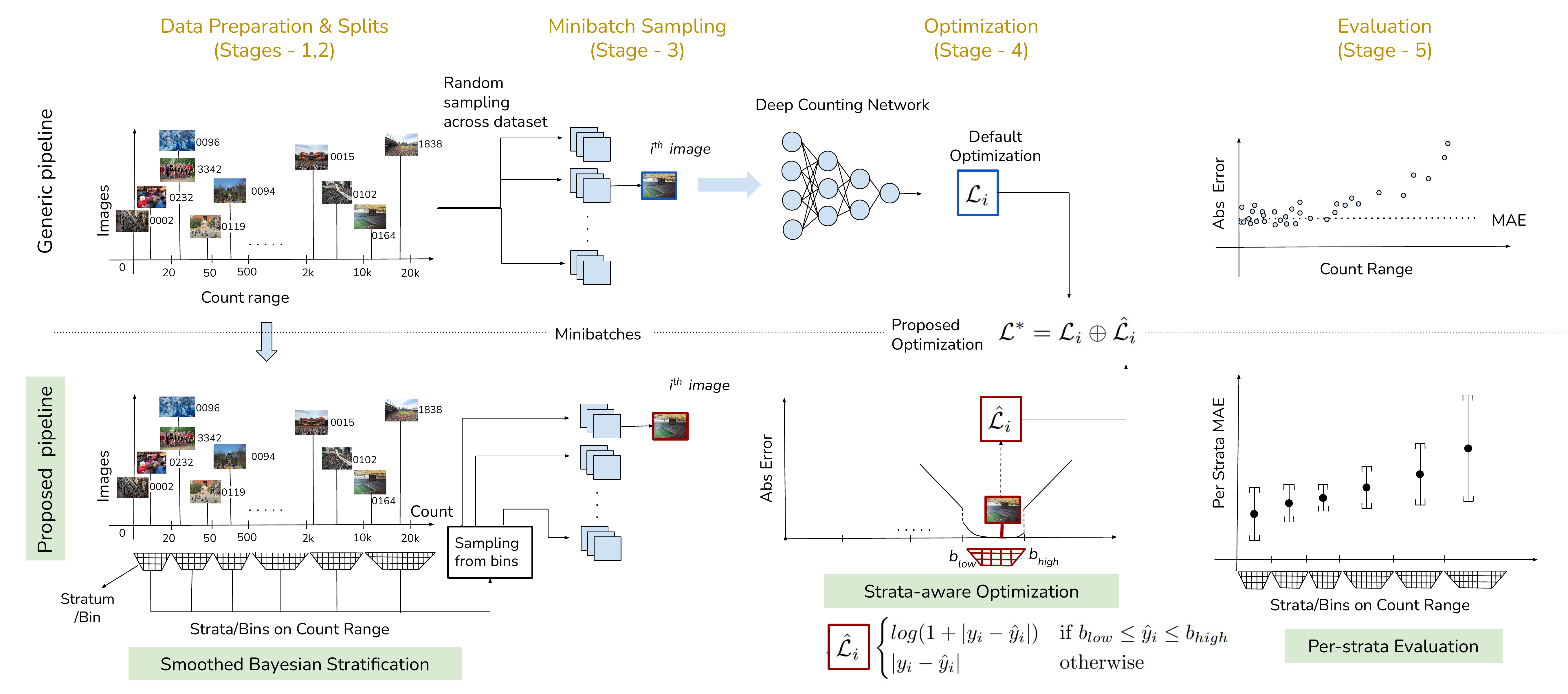}
\caption{An overview diagram depicting the generally employed processing pipeline of a crowd-counting approach (top) and the proposed modifications we introduce in this work (bottom). See Section~\ref{sec:proposedapproach} for details.} \label{fig:prototypicalAndChanges}
\end{teaserfigure}

\maketitle

\section{Introduction}
\label{section:intro}

Crowd counting is the technique of determining the number of people in a given image. Estimating count from images has significant applications in urban planning, surveillance in industries, hospitals and other establishments. Given an image, deep counting networks regress a single value representing the number of people in the image. Deep networks in crowd counting are typically trained on images and density maps generated from point annotations.

\begin{figure}[!t]
\centering
\includegraphics[width=0.5\textwidth]{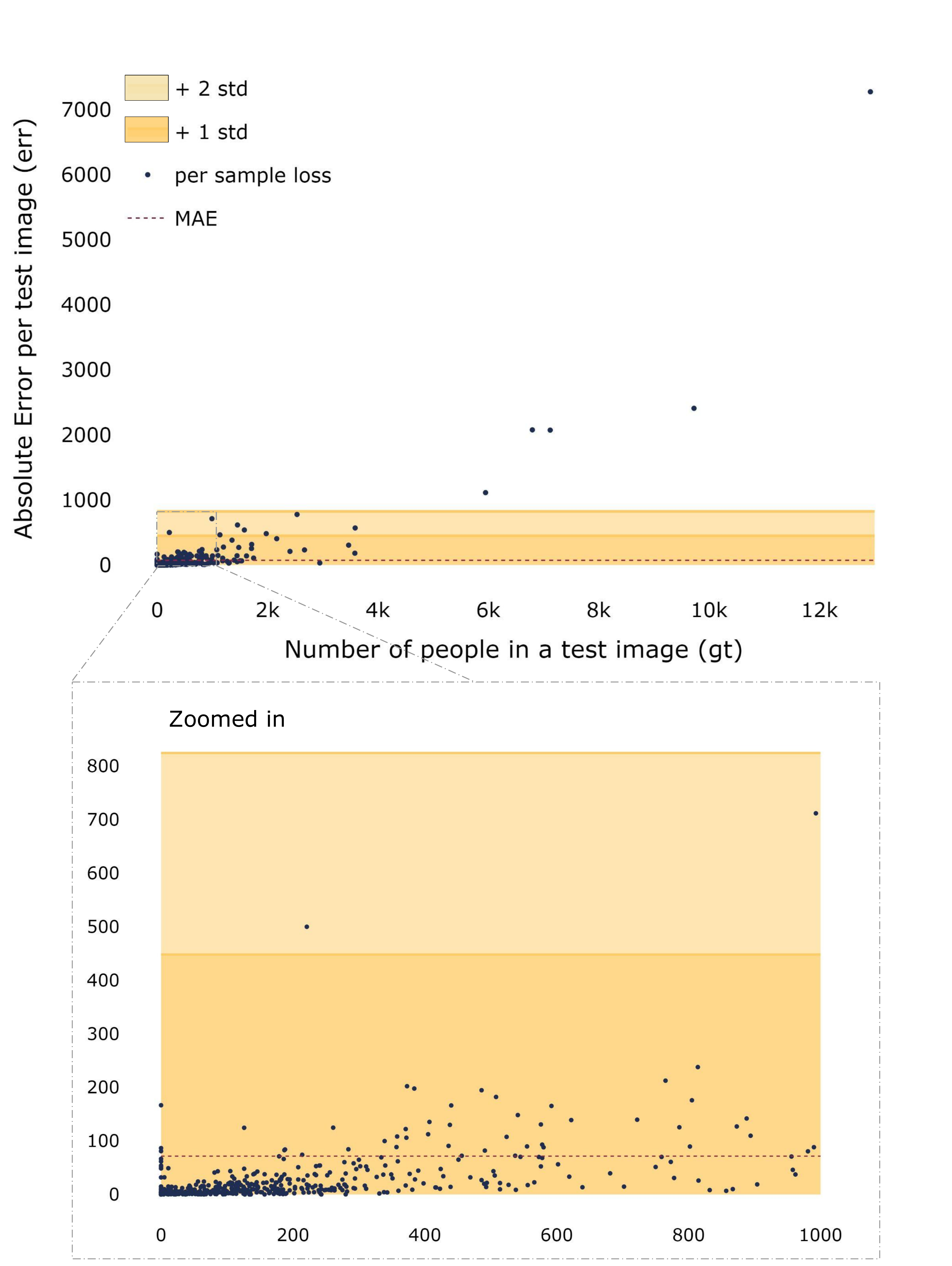}
\caption{The scatter plot of ground-truth counts and absolute errors by DM-Count~\cite{wang2020DMCount} on the NWPU dataset~\cite{gao2020nwpu}. The Mean Absolute Error (MAE) is $71.71$, but the standard deviation is multiple orders of magnitude larger: $376.40$. The zoomed in plot shows that even for lowest count (0 people), error is significantly larger than $0$. Clearly, MAE is a poor representative of performance across count range.}
\label{fig:1}
\end{figure}

Recent large-scale datasets used to train deep counting networks include Shanghai Tech~\cite{mcnn}, UCF-QNRF~\cite{idrees2018composition} and NWPU-Crowd~\cite{gao2020nwpu}. Although these datasets have considerably helped advance the state-of-the-art in crowd counting approaches, some issues remain to be addressed. A particularly alarming issue is the heavy-tailed and discontinuous distribution of crowd counts. Specifically, these datasets tend to contain a large number of images with small (people) count and a rather limited number of images with a large count (see Figure~\ref{fig:1}). 

The skew in the data distribution affects all aspects of the problem. It induces imbalance in minibatch sampling, optimization and evaluation. Since the default evaluation protocol (averaging over test errors) does not take the data distribution skew into account, the resulting score (e.g. Mean Absolute Error (MAE)) exhibits high standard deviation, often $2-3$ orders of magnitude higher than MAE itself (see Figure~\ref{fig:1}). This high deviation prevents mean score from being considered as a reliable performance statistic. Since error deviation is not reported in literature, this issue has gone unaddressed so far.

To address issues mentioned above, we propose an approach that actively factors in the count distribution and its skew at every stage of the problem (see Figure~\ref{fig:prototypicalAndChanges}). As the first step, we devise an algorithm for partitioning the count range into balanced strata (bins) using Bayesian optimality as a criterion (Sec.~\ref{sec:proposedapproach}). The balanced bins form the basis for minibatch sampling (Sec.~\ref{sec:stage3}). We also formulate a loss function that additionally penalizes error based on the ground-truth binning (Sec.~\ref{sec:stage4}). Instead of reporting a single performance summary statistic (MAE) across the entire test set range, we report bin-wise statistics and aggregate these statistics in a principled manner (Sec.~\ref{subsec:stage5}) to report the overall score. We perform comparative evaluation involving representative state-of-the-art deep counting networks~\cite{wang2020DMCount,ma2019bayesian,gao2019scar,xhp2019SDCNet,zhu2019dual}. 
Our results (Sec.~\ref{sec:results}) demonstrate that the proposed approach results in a noticeable reduction of error deviation compared to the default (no-binning) procedure. More generally, our approach helps both designers and end-users determine performance for various count ranges and select from among various approaches based on their relative performance within these ranges. 

Code, pretrained models and visualizations can be accessed from our project page \link{deepcount.iiit.ac.in}.

\section{Related Work}

To the best of our knowledge, no works have analyzed the processing pipeline for crowd counting in entirety. In this section, we review works which aim to address some aspects raised in the earlier section.

\noindent \textit{Density-based crowd counting:} Deep Convolutional Networks which represent the target count as a density map form the most popular class of approaches~\cite{mcnn,multiscale,csr,can}. Some approaches have attempted to address count distribution imbalance, although in an indirect manner. Sam \textit{et al.}~\cite{switch} propose a switching CNN based model which employs three regressors and a classifier which selects the best regressor to which an input patch is to be routed. There have also been attempts at reducing the skew at the patch level as in Xiong \textit{et al.}~\cite{xhp2019SDCNet}. They discretize the count range into a set of intervals and  design a classifier on these intervals, thereby converting an open set regression problem to a closed set classification one.

\noindent \textit{Point-based crowd counting:} To overcome the performance sensitivity to density map preparation, recent approaches use point annotations directly to estimate count. Ma \textit{et. al.}~\cite{ma2019bayesian} use a novel loss function that constructs a density distribution indirectly from the point annotations. Wang \textit{et al}.~\cite{wang2020DMCount} employ the optimal transport (OT) loss to find similarity between predicted density map values and ground truth binary point map and a total variation loss to stabilize the OT computation. 

\noindent \textit{Evaluation methods:} Mean Absolute Error (MAE) and Mean Squared Error (MSE) are the most prevalent evaluation measures in crowd counting approaches, with MAE usually being the more direct measure. More recently, some attempts have been made to examine MAE statistics based on percentage errors, illumination levels and scene levels to characterize performance~\cite{gao2020nwpu}. However, these are post-hoc measures and do not tackle imbalance which crops up in other stages of the standard pipeline employed for crowd counting.

\section{Proposed method}
\label{sec:proposedapproach}

\subsection{Standard Processing Pipeline}
\label{sec:pipe}
As depicted in Figure~\ref{fig:prototypicalAndChanges}, any standard approach to crowd-counting can be considered to have five stages:

\begin{itemize}[noitemsep]
    \item \textit{Stage-1 (Data preparation):} In this stage, images and corresponding counts are processed suitably and are provided as input and output to a reference deep network. This stage includes standard procedures such as image cropping and resizing, density map preparation, {\em  etc.}
    \item \textit{Stage-2 (Creating data splits):} The prepared data is partitioned into training, validation and test splits according to a pre-defined split ratio ({\em e.g.}, $65\%,15\%,20\%$). 
    \item \textit{Stage-3 (Minibatch creation):} The deep network is trained using a subset of data randomly sampled from the training set, usually referred as a minibatch. The training set is partitioned into minibatches for each training epoch.
    \item \textit{Stage-4 (Optimization):} The parameters of the deep network are optimized for a loss function at the minibatch level. 
    \item \textit{Stage-5 (Evaluation):} A standard performance measure ({\em e.g.}, MAE) is used for evaluating the model on the validation or the test set. 
\end{itemize}

Each of these stages involves a set of assumptions which are often implicit. For instance, the train-validation-test splitting (Stage-2) and minibatch creation (Stage-3) assume that the distribution over the targets (counts) is uniform. However, target distributions for standard crowd counting datasets are heavy-tailed. Due to the uniform nature of sampling, the data splits and consequently, the training minibatches, exhibit the same heavy-tailed distribution. This skew induces a bias which penalizes samples in the tail during optimization (Stage-4). Due to this bias, the statistical summary measures ({\em e.g.}, MSE, MAE) fail as representative measures of performance (Stage-5). 

To address these issues, we revisit the entire problem setting and propose alternative paradigms for the stages mentioned previously. We leave Stage-1 untouched and describe our modifications to the subsequent stages. 

\subsection{Revisiting Stage 2 (Creating Data splits)}
\label{subsec:stage2}

As mentioned earlier, the standard sampling procedure for creating train-validation-test splits implicitly assumes a uniform distribution over the target range. However, doing so causes the tail portion of the distribution to be under-represented. A fundamental reason for this effect is that the sampling is conducted at too fine a resolution, {\em i.e.} at the level of individual counts. 

One approach to address this issue is to coarsen the resolution and partition the count range into bins (strata) that are optimal for uniform sampling. Formally, let the total number of images be $N$ and suppose the count range over the data samples is $R=[0,C]$, where $C$ is the maximum crowd count. The count data $\Dcal$ can be represented in terms of observed discrete counts $c_i$ and their frequencies $f_i$, as $\Dcal = \left \{ \langle c_i, f_i \rangle \left|\  i = 1,...m \right. \right\}$, where $m$ is total number of distinct counts in the dataset. Thus, $c_1=  0, c_m = C$. Consider a partitioning of the counts into $N_b$ bins as:

\begin{equation}
\label{eq:1}
    \Pscr(1,N) \equiv \lbrace [n_{k-1},n_k-1] \rbrace, k = 1,2,3\ldots N_b
\end{equation}

where $n_{k-1}$ represents the start index of the $k^{th}$ bin. Note that $n_0=0$ and $n_{N_b}-1 = C$. For simplicity, we drop the reference to $(1,N)$ when referring to $\Pscr(1,N)$ in what follows. 

\subsubsection{Partition Prior}
\label{sec:partitionprior}

We formulate the prior over partitions in terms of number of bins $N_b$ in a candidate partition. In what follows, we refer to this prior distribution as $P(N_{b})$. To avoid the degenerate case in which each unique count in the range might land up in its own bin, we impose constraints over the number of bins~\cite{Scargle2013}. Specifically, we use a geometric prior to assign lower probability to a partition containing larger bin counts:

\begin{align}
\label{eq:3}
    P(N_b;\gamma) =  
    \begin{cases}
    P_0 \; \gamma^{N_b} & \text{if } 1 \leqslant N_b \leqslant \alpha \\
    0 & \text{otherwise}
  \end{cases}
\end{align}

where $P_0$ is a normalization constant. $\gamma < 1$ is a parameter which affects the distribution profile and $\alpha$ controls the practical effectiveness of the upper bound on $N_b$. Applying the laws of probability to $P(N_b)$ and solving for $P_0$, we  obtain: 

\begin{equation}
\label{eqn:prior}
    P(N_{b};\gamma) = \frac{1 - \gamma}{1 - \gamma^{\alpha}} \gamma^{N_{b}}
\end{equation}

\subsubsection{Partition Likelihood}
\label{sec:partitionlkhood}

\begin{figure}[!t]
\centering
\includegraphics[width=0.3548\textwidth]{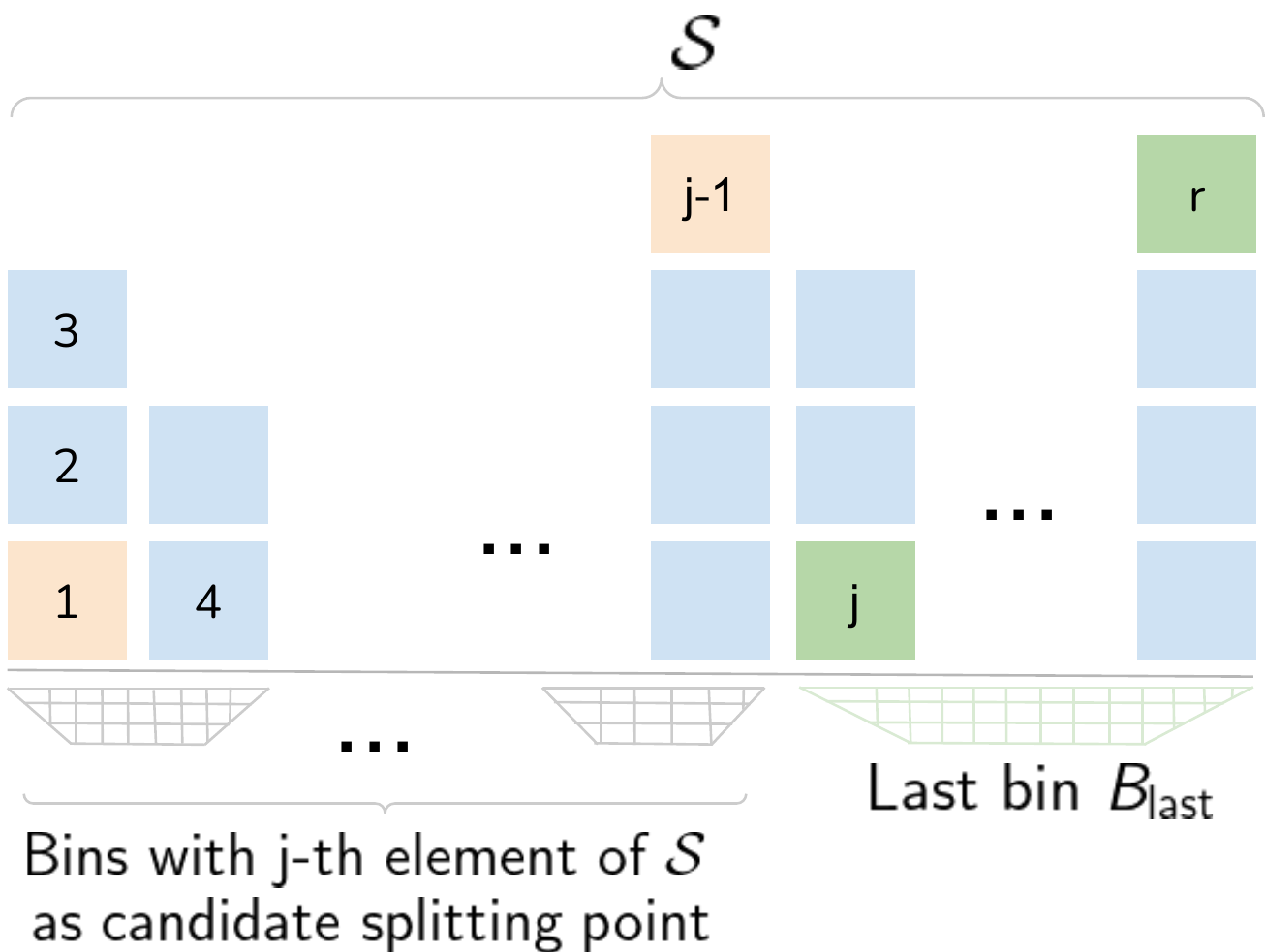}
\caption{A candidate partitioning of a subsequence of $\Scal$ ending with the $r^{th}$ element of $\Scal$. Finding the optimal partitioning can be thought of as a search over such candidate partitions. Refer to Sec.~\ref{sec:optimalpartitioning}.}
\label{fig:candidatepartition}
\end{figure}

The likelihood for a partition $\Pscr$ is defined in terms of the likelihood of each constituent bin in the partition. Let $m_k$ be the width of bin $B_k$. Let the count frequencies of the $m_k$ distinct counts within the bin be denoted by $x_1,x_2,\ldots x_{m_k}$ respectively. We model the likelihood for each bin as a multinomial distribution:

\begin{equation} \label{eqn:binlkhood}
\begin{split}
lik(B_k) & = lik(x_1,\dots , x_{m_k} ; p_1, \dots, p_{m_k}) \\
 & = \frac{X_k!}{x_1! x_2! \ldots x_{m_k}!} \prod_{j=1}^{m_k} p_{j} ^{x_j}
\end{split}
\end{equation}

where $X_k = \sum_{j=1}^{m_k} x_j$ and $p_{j}$ is probability of the $j^{th}$ count. Assuming bin-level independence, the \textit{log} likelihood of the partition can be expressed as:

\begin{equation}
\label{eq:7}
    lik[\Pscr] = \sum_{k=1}^{N_b} lik(B_k)
\end{equation}

\subsubsection{Optimal Partitioning}
\label{sec:optimalpartitioning}

Given the count range $R = [0,C]$, at one extreme, we can have a partitioning wherein all data lies in a single bin. At the other extreme, we can have a partitioning wherein each unique integer in the range $R$ is a bin. Thus, finding the optimal partitioning can be thought of as a search over candidate partitions that lie between these two extremes. 

To solve this task efficiently, we adopt a dynamic programming approach~\cite{Scargle2013}. To begin with, we transform the count frequency data $\Dcal$ into a sequence of counts $c_1, c_2....c_m$ where $c_i$ is repeated $f_i$ times, {\em i.e.}, $\Scal : \lbrace c_i,c_i,\ldots (f_i $ times$, 1 \leqslant i \leqslant m) \rbrace$.  Let $\Fcal_{opt}(1,r)$ be the optimal Maximum A Potseriori (MAP) score for the partitioning of a subsequence of $\Scal$ ending with the $r^{th}$ element of $\Scal$. Following the principle of optimality, we have:

\begin{equation}
\begin{aligned}
    \Fcal_{opt}(1,r) =  
    \begin{cases}
    0, \text{ if }  r =1 \\
   \underset{1 < j \leqslant r}{\text{max}} \Big[ \texttt{ best(1,j-1)}  + lik(B_{last})(j,r) \\ \;\;\;\;\;\;\;\;\;\; + \; log \: P(b_j;\gamma) \Big] \text{ if r=2,3\ldots N} 
  \end{cases}
\end{aligned}    
\label{eqn:fopt}
\end{equation}

where \texttt{best(1,j-1)} is the memoized (precomputed and stored) best likelihood value (Eqn.~\ref{eq:7}) for the sub-partition ending in the $(j-1)^{th}$ element, $lik(B_{last})(j,r)$ is the likelihood of the final bin containing the subsequence beginning at the $\Scal$'s $j^{th}$ element and ending with the $r^{th}$ element (see Fig.~\ref{fig:candidatepartition}).  $log \: P(b_j;\gamma)$ is the prior on number of bins (Eqn.~\ref{eqn:prior}). More concretely, $b_j$ is the number of bins that form with $\Scal$'s $j^{th}$ element as the split location for the last bin.  

Note that the MAP formulation of $\Fcal_{opt}(1,r)$ incorporates the partition likelihood and prior in a Bayesian manner. With respect to the formulation in Eqn~\ref{eqn:fopt}, the optimal set of bins corresponds to the ones obtained for $\Fcal_{opt}(1,|\Scal|)$,  where $|\Scal|$ is the number of elements in sequence $\Scal$.

\begin{algorithm}[!t]
  \caption{Optimal Bins}
  \label{algorithm:1}
  \begin{algorithmic}[1]
  \Procedure{OptimalBins}{$\Dcal$}
\LineComment{\textbf{Input} data $\Dcal$ }
\LineComment{\textbf{Output} Optimal bins $bins_{best}$}
\LineComment{\textcolor{blue}{Grid search values for $\gamma$ (Sec.~\ref{sec:partitionprior})}}
 \State $\Gamma = [0.1,0.2,\ldots0.9]$
\LineComment{\textcolor{blue}{Grid search values for train-test ratios}}
 \State $ratios = [0.1,0.2,0.25]$
\LineComment{\textcolor{blue}{Cross-validation repeat factor}}
 \State $seeds= 10$

 \For {$\gamma \text{ in } \Gamma$}
     \For{$r \text{ in } ratios$}
         \For{$f \text{ in } [0:1:seeds]$}
                \State $\Dcal_{f}$ = shuffle($\Dcal$,$seed=f$);
                \LineComment{\textcolor{blue}{Algorithm~\ref{algorithm:2}}}
                \State $lik_{f,r,\gamma}$ = \Call{FindLikelihood}{$\Dcal_{f},r,\gamma$}
         \EndFor
        \LineComment{\textcolor{blue}{Compute average likelihood for a fixed $\gamma$ and $r$}}
         \State $lik_{r,\gamma} =$ \Call{Mean}{$lik_{f,r,\gamma}$}
    \EndFor
\EndFor
\LineComment{\textcolor{blue}{To find the best $\gamma$ across all $ratios$,}}
\LineComment{\textcolor{blue}{descending sort by likelihood for each ratio $r$.}}
\LineComment{\textcolor{blue}{For each $\gamma$, sum indices of corresponding location}}
\LineComment{\textcolor{blue}{in sorted order of earlier step.}}
\For {$\gamma \text{ in } \Gamma$}
\State $idxsum_{\gamma} = 0$
\For{$r \text{ in } ratios$}
\State $idxsum_{\gamma} += $ \Call{GetDescendingIndx}{$lik_{r,\gamma}$}
\EndFor
\EndFor
\LineComment{\textcolor{blue}{The best $\gamma$ is one with lowest index sum.}}
\State $\gamma_{best}$ = $\underset{\gamma}{\text{ argmin }} idxsum_{\gamma}$
\LineComment{\textcolor{blue}{Use the best $\gamma$ and determine optimal partitions (Sec.~\ref{subsec:stage2}).}}
\State $bins_{best}$ =  \Call{BayesianOptimalBins}{$\Dcal,prior=\gamma_{best}$}
  \EndProcedure
\end{algorithmic}  
\end{algorithm}

\begin{algorithm}[!t]
\caption{Algorithm to find likelihood of a held out subset}
\begin{algorithmic}[2]
\Procedure{FindLikelihood}{$\Dcal,ratio,\gamma$}
\LineComment{\textbf{Input} Data $\Dcal$,train-test split ratio $ratio$, prior param $\gamma$ }
\LineComment{\textbf{Output} Likelihood $lik$ of $\Dcal$'s test subset}
\LineComment{\textcolor{blue}{Split data into train, test as per $ratio$}}
\State $train$ $,test$ $=$ \Call{SplitData}{$\Dcal,ratio$}
\LineComment{\textcolor{blue}{Find optimal bins using train set} (Sec.~\ref{subsec:stage2})} 
\State $bins=$ \Call{BayesianOptimalBins}{$train,prior=\gamma $}
\LineComment{\textcolor{blue}{Find likelihood of test set}}
\LineComment{\textcolor{blue}{wrt optimal bins found earlier (Sec.~\ref{sec:partitionlkhood})}}
\State $lik$ $=$ \Call{ComputeBinsLkhood}{$test,bins$}
\EndProcedure
\end{algorithmic}
\label{algorithm:2}
\end{algorithm}

\subsubsection{Additive Smoothing}
\label{sec:additivesmoothing}

The sample distribution in crowd datasets is not only heavy tailed, but also sparse at the tail end. In other words, the distribution is characterized by large count spans which do not have any sample associated with them. This causes the binning procedure described in this section to output a large number of sparsely filled bins. To mitigate this effect, we perform additive smoothing~\cite{daniel} on the data before binning. Formally, a smoothing factor $\beta$ is added to each distinct count across the count range $R=[0,C]$. In our case, $\beta=1$.

\subsubsection{Grid-search for optimal hyperparameters}
\label{sub:modbay}

To determine the optimal set of bins, we first perform a grid search with cross-validation over a range of values for (i) distribution profile parameter $\gamma$ (Eqn.~\ref{eq:3}) (ii) the train-validation split ratios. Having determined the optimal hyperparameter $\gamma_{best}$, we utilize the same to obtain the optimal set of bins, as outlined in Algorithm ~\ref{algorithm:1}. 
\subsection{Revisiting Stage 3: Minibatch Creation}
\label{sec:stage3}

To address the skew induced by the heavy-tailed, discontinuous count distribution of data samples, we bin the data optimally using the procedure described in Section~\ref{subsec:stage2}. To populate a minibatch using our Round Robin (RR) method, we pick a data sample randomly from each of the bins in a round robin fashion, beginning at the first bin. This process is repeated until all the bins have been selected or the minibatch is full. We continue this process until the entire training dataset is accounted for as an epoch ({\em i.e.}, in terms of minibatches). This procedure is followed for each epoch. 

Another variant of binning which we consider is Random Sampling (RS) procedure where a bin is first picked randomly from available bins and a data sample is picked randomly from the randomly selected bin. A procedure similar to Round Robin (RR) is used to populate an epoch's equivalent of training data. Effectively, both our procedures ensure that the mini-batches are balanced in terms of their count range unlike the standard random shuffle-based approach. We analyze the results on both the binning strategies during evaluation (Sec.~\ref{sec:results}).

\subsection{Revisiting Stage 4: Optimization}
\label{sec:stage4}

The standard protocol for optimizing a deep counting network is to minimize the per-instance loss averaged over the minibatch. However, one is confronted with the same issues (imbalance, bias) as those faced during minibatch creation (Sec.~\ref{sec:stage3}). As a consequence, the trained networks exhibit high variance for the error term $|y - \hat{y}|$, where $y$ is the ground-truth count and $\hat{y}$ is the predicted count.

\begin{figure}[!t]
\centering
\includegraphics[width=0.45\textwidth]{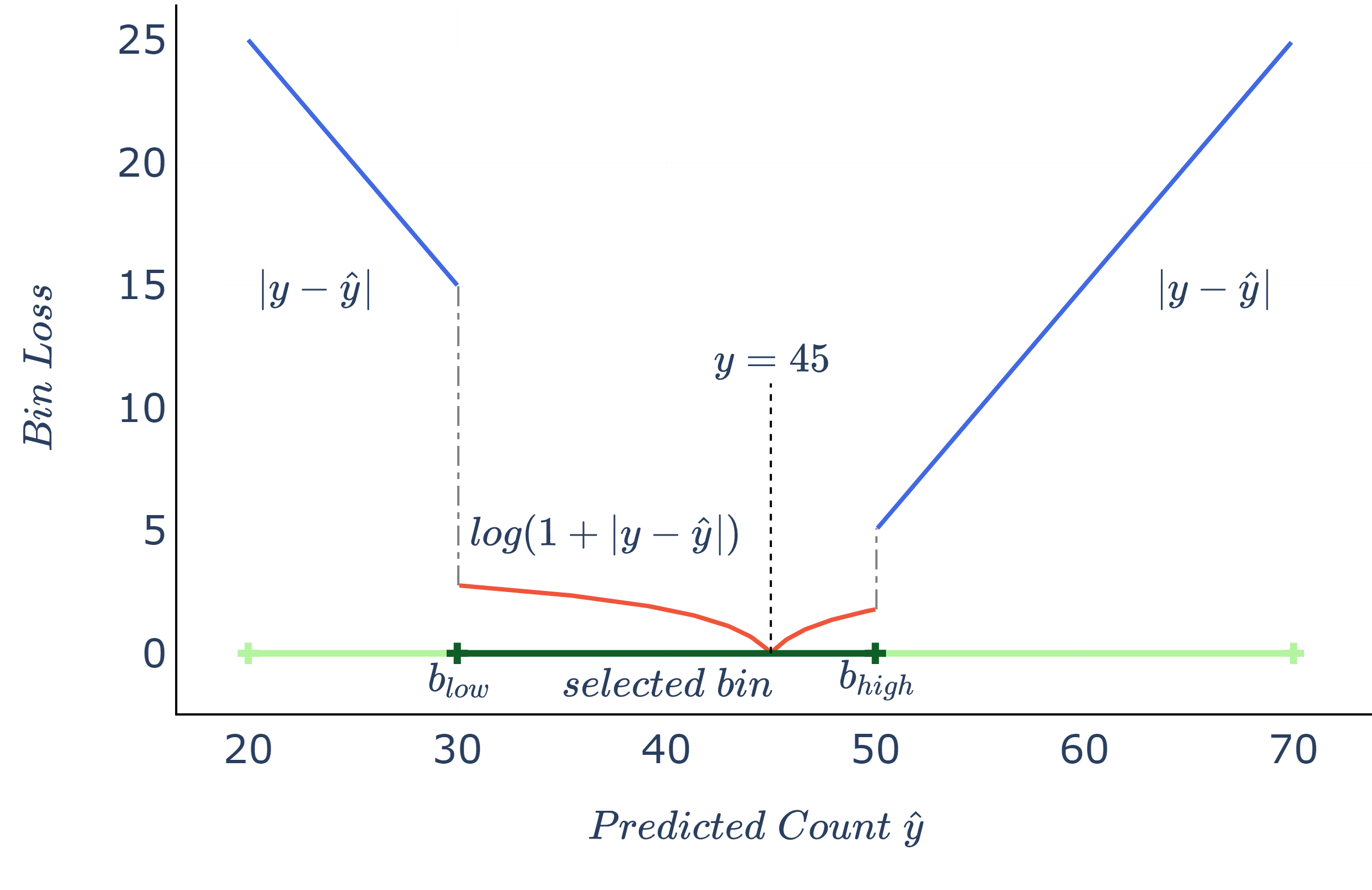}
\caption{Bin Loss Function : The figure depicts the ground truth count $y=45$ and the loss function variation with respect to the predicted count $\hat{y}$ inside the bin ($log(1+ |y - \hat{y}|)$) and outside ($ |y - \hat{y}|$). The reference bin is highlighted in dark green. Refer to Sec.~\ref{sec:stage4} for details.}
\label{fig:3}
\end{figure}

To enable data-distribution aware optimization, we introduce a novel bin sensitive loss function $\widehat{\mathcal{L}}$. Instead of the loss depending solely on the error, we also consider the count bin to which the data sample belongs and whether the predicted count $\hat{y}$ lies within this bin or outside it. If $\hat{y}$ lies within the bin, we impose a smaller logarithmic penalty.  If the count value lies outside, we impose a linear penalty. Formally, our strata-aware loss function is defined as:  

\begin{equation}
  \widehat{\mathcal{L}} =   \begin{cases}\label{eq:8}
  \lambda_1 \; log(1+ |y - \hat{y}|) & \text{if $ b_{low} \leqslant  \hat{y}  \leqslant b_{high} $ } \\
  |y - \hat{y}| & \text{otherwise}
\end{cases} 
\end{equation}

where $b_{low}$ and $b_{high}$ are defined by the bin that $y$ belongs to (see Fig.~\ref{fig:3}) and $\lambda_1$ is a weighting factor of the log component. This loss is added as an additive component to the default model loss to encourage strata-aware optimization.

\subsection{Revisiting Stage 5: Evaluation}
\label{subsec:stage5}

The discontinuous and heavy-tailed distribution of samples affects the evaluation stage as well. Coupled with lack of bin-level awareness during optimization, an outlier effect arises which causes the default measures ({\em e.g.}, MSE, MAE) to be ineffective representatives of performance \textit{across} the entire count range. Even more worryingly, the standard deviation of error tends to be at the same level as the mean statistic. Instead of using a single pair of numbers (mean, standard deviation) to characterize performance across the entire count range, we make the following proposals.

\textit{One,} the evaluation measure must be reported at the level of each bin. This provides a more comprehensive picture of performance. Additionally, it also helps compare the relative effectiveness of various counting networks for smaller and larger counts. \textit{Two,} even if an overall summary statistic over the test set is deemed necessary, the mean and standard deviation of bin-level performance measures are combined in a statistically sound manner. Let the mean and standard deviations for the individual bins be $(\mu_i,\sigma_i); i=1,2,\ldots N_b$ and let the number of samples in each bin be $n_i$. We compute the pooled mean and standard deviation as their weighted average:
    
\begin{equation}
    \mu_{pool} =\frac{n_1 \mu_1 +  n_2 \mu_2 + \ldots + n_{N_b} \mu_{N_b} }{n_1 + n_2 + \ldots + n_{N_b}}
\end{equation}

\begin{equation}
    \sigma^2_{pool} =\frac{n_1 \sigma^2_1 +  n_2 \sigma^2_2 + \ldots + n_{N_b} \sigma^2_{N_b} }{n_1 + n_2 + \ldots + n_{N_b}} 
\end{equation}

\section{Experimental Setup}
\label{sec:experiments}

\begin{figure}[!t]
\centering
\includegraphics[width=0.5\textwidth]{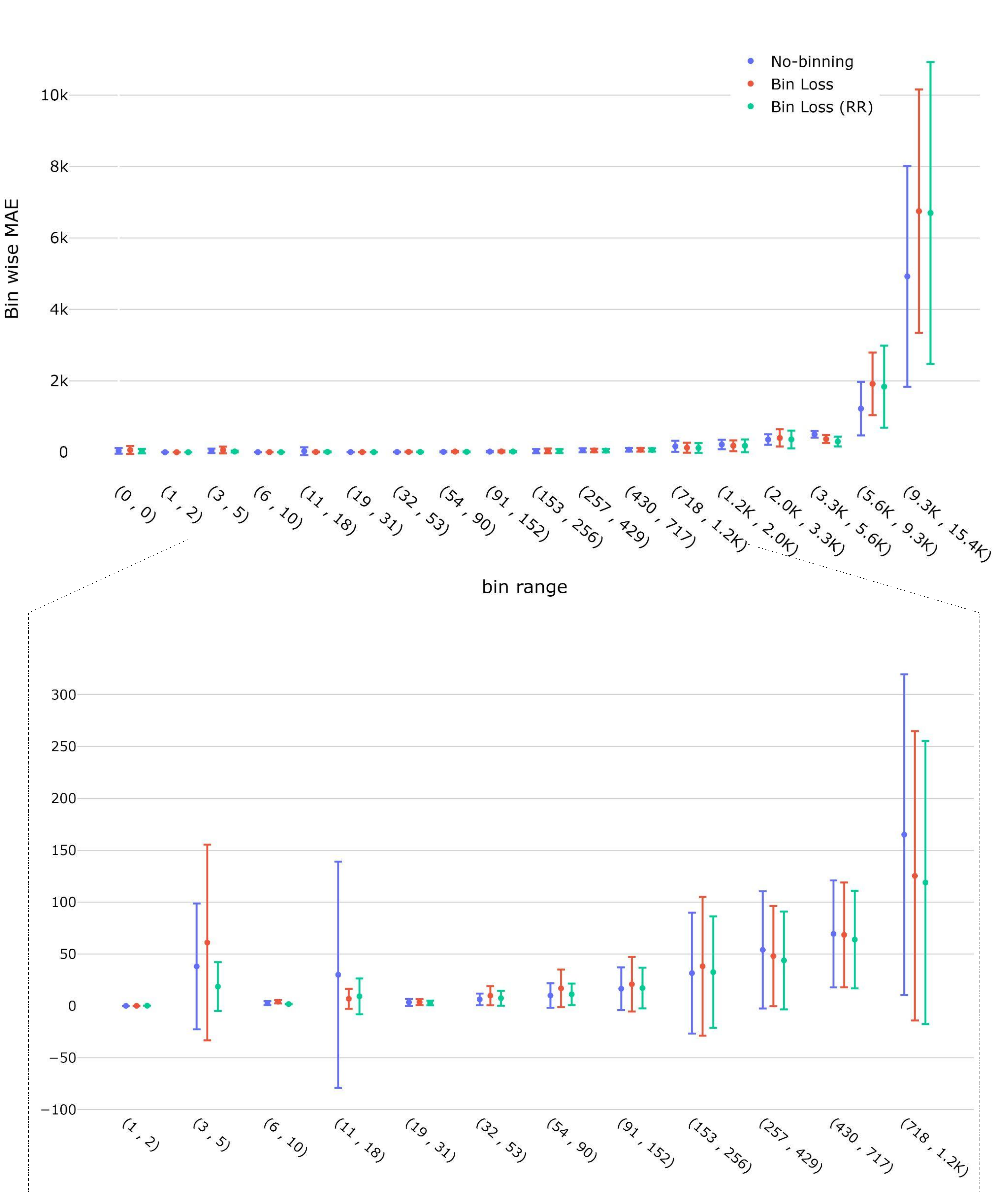}

\caption{Per-bin performance of DM-Count~\cite{wang2020DMCount} on NWPU dataset~\cite{gao2020nwpu} for different binning schemes (color-coded). MAE is represented by a dot and error bars represent standard deviation. Bins in range $[1,1.2k]$ are shown zoomed in for better visibility. The comparatively larger deviations for the no-binning scheme are clearly evident.}
\label{fig:5}
\end{figure}

We perform experiments with two large-scale crowd counting datasets NWPU~\cite{gao2020nwpu} and UCF-QNRF~\cite{idrees2018composition}  as well as two variants of the medium-scale dataset ShanghaiTech(A,B)~\cite{mcnn}. Although we revisit all stages of the problem pipeline, we retain the standard train and test datasets for consistency. To determine optimal bin hyperparameters (Section~\ref{subsec:stage2}), we isolate a random $20\%$ subset of the train set and use the same for validation. Since NWPU's test set is not directly available, we use the publicly available validation set as the test set and report results on the same. We also compare the two different binning schemes mentioned in Section.~\ref{sec:stage3}, {\it viz.}, round-robin (RR) and random selection (RS).
For evaluation, we utilize representative and recent state-of-the-art crowd counting networks, {\em viz.}, DM-Count~\cite{wang2020DMCount}, Bayesian Crowd Counting (BL)~\cite{ma2019bayesian}, SCAR~\cite{gao2019scar}, SFA-Net~\cite{zhu2019dual}, S-DCNet~\cite{xhp2019SDCNet}. These papers report results on the ShanghaiTech  and UCF-QNRF datasets but not on NWPU (except for DM-Count). Therefore, we report respective test set results by training these networks on the NWPU dataset as well. 

The network architecture, ground truth generation, augmentation and image pre-processing steps are used as mentioned in the respective works. We use the hyperparameters, optimizers and loss functions used as suggested in the original implementations of the networks. As mentioned previously, we add the bin-aware loss function (Sec.~\ref{sec:stage4}) to the original loss function used by the models during optimization. We compute the per-bin MAE and associated standard deviation. We also aggregate the resulting statistics to obtain an overall performance score across the bins (Sec.~\ref{subsec:stage5}). Although not directly comparable to our proposed performance score, we also report the standard MAE (which does not involve any binning) as computed by existing works. As a new addition, we also report the error's standard deviation. For baseline comparison, we also train models using the default (no-binning) procedure and without the bin-aware loss function included. 

\section{Results}
\label{sec:results}

\subsection{Bin-level results}

\begin{figure}[!t]
\centering
\includegraphics[width=0.5\textwidth]{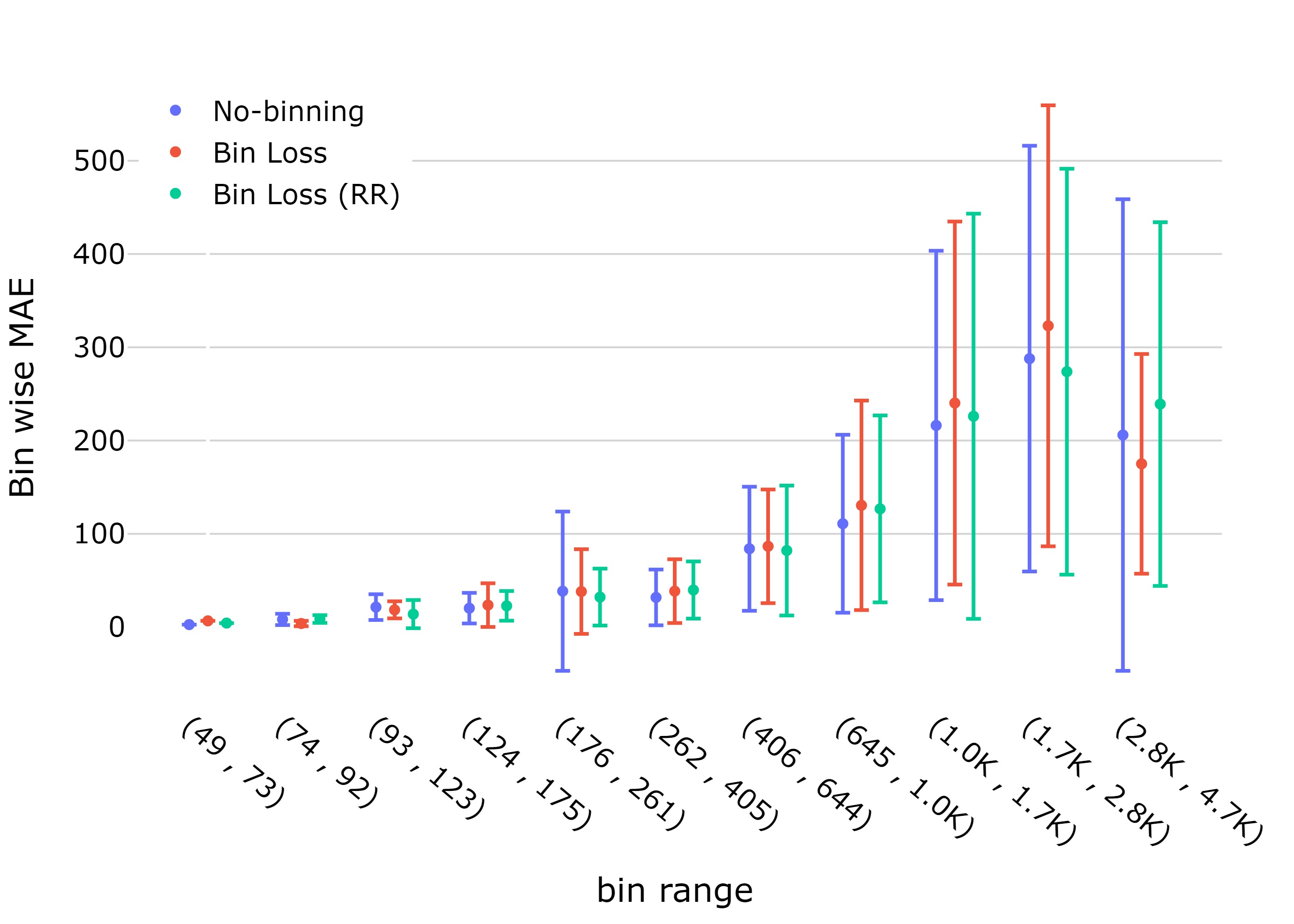}
\caption{Per-bin performance of DM-Count on UCF dataset. The comparatively larger deviations for no-binning scheme are clearly evident as with other plots.}
\label{fig:6}
\end{figure}

\begin{figure}[!t]
\centering
\includegraphics[width=0.5\textwidth]{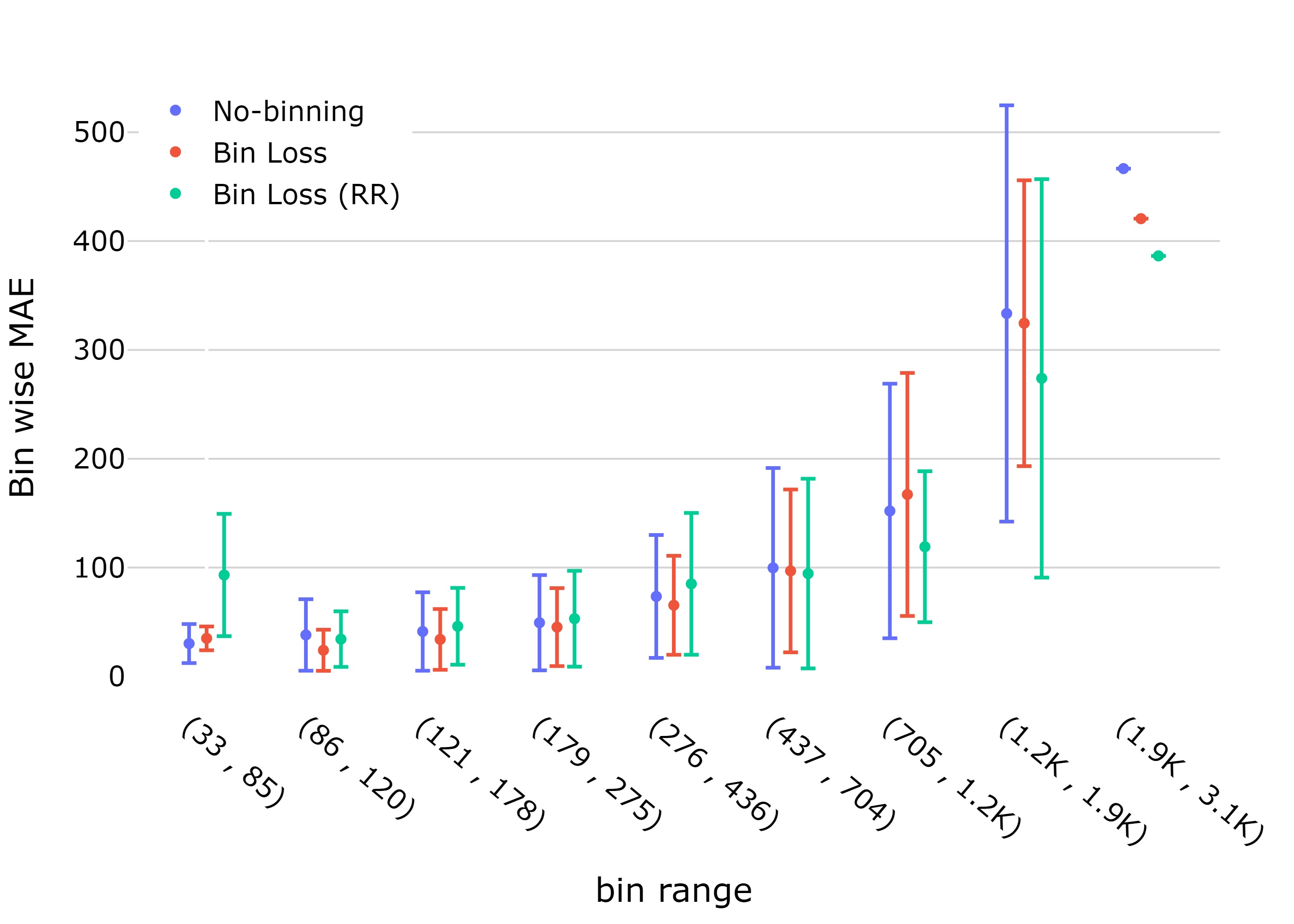}
\caption{Per-bin performance of DM-Count on STA dataset. Similar to our observation in the earlier plots, the comparatively larger deviations for the no-binning scheme are clearly evident.}
\label{fig:7}
\end{figure}

\begin{figure}[!t]
\centering
\includegraphics[width=0.5\textwidth]{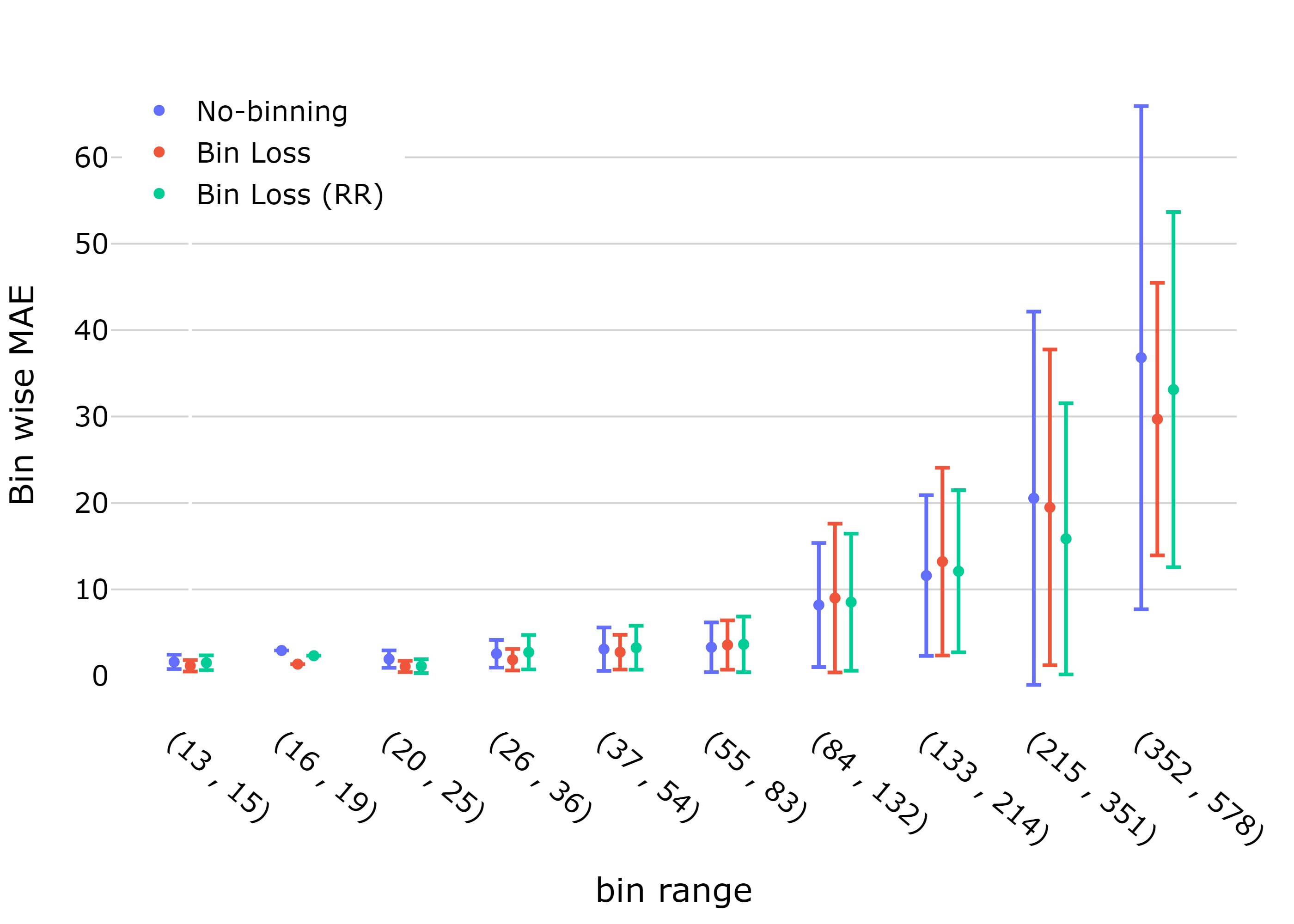}
\caption{Per-bin performance of DM-Count on STB dataset. The comparatively larger deviations for the no-binning scheme are clearly evident, like in the earlier plots.}
\label{fig:8}
\end{figure}

The bin-level mean error scores and the corresponding standard deviation bars can be viewed for a selection of different datasets and binning schemes in Figures~\ref{fig:5},~\ref{fig:6},~\ref{fig:7} and~\ref{fig:8}. The comparatively large deviations typically incurred when binning is not used can clearly be seen. Also note that the bin-level plots provide a larger perspective on the performance of the approach across the count range, in contrast to a single number which is usually reported. Our project page \link{deepcount.iiit.ac.in} contains interactive visualizations for examining results on a per-dataset and per-model (approach) basis.

\subsection{Aggregate results}

The aggregate scores (described in Section~\ref{subsec:stage5}) can be viewed in Table~\ref{table:results} -- refer to the three gray-shaded columns. Across networks and datasets, a reduction in error standard deviation is clearly apparent when bin-aware loss is used (relative to the no-binning counterpart). The aggregate scores reinforce the trend seen in the bin-level plots discussed previously. The reduction in standard deviation compensates for the marginally inferior mean score (compared to no-binning) in some cases. As the blue highlighted results in Table~\ref{table:results} indicate, binning schemes provide the best overall aggregate results across the datasets (except for the smaller count STB dataset).   

In the last column of Table~\ref{table:results}, we also present the usually reported MAE measure. The results using models made available by authors are indicated. For the first time, we also report the standard deviation for the sake of completeness and consistency. Note that the numbers in this column are not directly comparable with other (gray) columns of the table due to the significant differences across the processing pipeline stages. However, the magnitude of the deviation incurred even by the state of the art approaches is somewhat alarming. It is also interesting to note that the MAE performance ranking for different networks differs significantly from the binning (Pooled MAE) results. For instance, BL~\cite{ma2019bayesian} is the best performer on UCF with Pooled MAE. A similar trend can be seen for the STA and STB datasets as well. Due to unavailability of BL-specific settings for NWPU dataset, we used the settings used for BL with UCF-QNRF. These settings may be sub-optimal and might be the reason BL underperforms on NWPU. 

In our experiments, we tried two minibatching schemes (balanced, random) to determine their effect on performance, if any (Section \ref{sec:stage3}). The aggregate results across datasets suggests that random sampling has better overall performance approximately half the time (Table~\ref{table:results}, first two columns). Also, the results suggest that random sampling of bins works best for top performing networks (DM-Count~\cite{wang2020DMCount}, BL~\cite{ma2019bayesian}) most of the time. 

\begin{table*}[!t]

\captionof{table}{Evaluation results on four benchmark datasets NWPU, UCF-QNRF, ShanghaiTech-A,B (STA,STB)  using the evaluation procedure in Sec.~\ref{subsec:stage5} on diverse models. The size of test set is indicated below dataset name. The columns represent minibatching schemes (Bin Loss: random bin selection (RS), Bin Loss(RR): round robin bin selection, No-binning: default procedure without binning). For each result, superscript denotes the standard deviation. The best result for each dataset is highlighted in blue. The best MAE and standard deviation of the absolute errors are highlighted in bold for each network. Note that gray highlighted columns of the table (Pooled MAE and standard deviation) are not directly comparable to the Global MAE and standard deviation values.}
\label{table:results}
\centering
\resizebox{0.9\linewidth}{!}
{
\centering
\begin{tabular}{c|c|r|aaa || c}
      & & & \multicolumn{3}{c||}{Pooled MAE and std} & \multicolumn{1}{c}{Global MAE and std} \\
\toprule
Size of Dataset & Dataset & Model  & Bin loss & Bin loss (RR) & No-binning & Pretrained,No-binning \\

\toprule

\multirow{10}{*}[-0.25em]{\rule{0pt}{2ex} Large }
& \multirow{5}{*}[-0.25em]{\rule{0pt}{2ex} ${ \text{\normalsize{NWPU}} \atop \mathbf{500}}$  } 
 & DM-Count \cite{wang2020DMCount} & $\;\:88.1^{\pm 236.7 } $ & \cellcolor{SkyBlue}$\;\:\mathbf{76.7}^{\pm \mathbf{205.0}}$ & $\;\:77.8^{\pm 214.9 }$ & $\;\:71.7 ^{\pm \:\:376.4}$ \cite{wang2020DMCount}  \\
& & BL \cite{ma2019bayesian}  & $112.9^{\pm 333.7 }$ & $114.8^{\pm \mathbf{320.3} }$ & $\mathbf{102.5}^{\pm 348.2 }$ & $ 102.5 ^{\pm 560.6 } $   \\
& &  S-DCNet \cite{xhp2019SDCNet} &  $213.4^{\pm 231.0}$  & $224.1^{\pm \mathbf{230.1} }$  & $ \mathbf{210.0}^{\pm 303.1 }$ & $ 248.7 ^{\pm 1161.9 } $ \\
& & SCAR \cite{gao2019scar}  & $112.8^{\pm \mathbf{321.3}}$ & $111.9^{\pm 325.6 }$  & $\mathbf{111.3}^{\pm 332.1 }$ & $ 111.3 ^{\pm 555.8 } $  \\
& & SFA-Net \cite{zhu2019dual}  & $136.0^{\pm 299.1 }$ & $\mathbf{116.4}^{\pm \mathbf{285.2} }$ & $125.0^{\pm 343.0 }$  & $ 163.4 ^{\pm 1072.1 } $ \\

\cmidrule(lr){2-7}

& \multirow{5}{*}[-0.25em]{\rule{0pt}{2ex} $\text{\normalsize{UCF}}\atop \mathbf{334}$} 
& DM-Count \cite{wang2020DMCount} & $103.8^{\pm \mathbf{107.5} }$ & $97.9^{\pm 109.1 }$ & $\mathbf{94.5}^{\pm 111.6 }$ & $ 85.9^{\pm 120.6} $ \cite{wang2020DMCount} \\
& & BL \cite{ma2019bayesian}  & \cellcolor{SkyBlue}$\;\:\mathbf{91.1}^{\pm \mathbf{100.3 }}$ & $ 92.1^{\pm 105.8 }$ & $98.3^{\pm 134.2 }$  & $ 87.1^{\pm 126.8} $ \cite{ma2019bayesian} \\
& &  S-DCNet \cite{xhp2019SDCNet} & $205.9^{\pm \mathbf{157.8} }$ & $\mathbf{199.2}^{\pm 164.8 }$ & $215.2^{\pm 190.0 }$ & $ 214.7 ^{\pm 277.7 } $  \\
& & SCAR \cite{gao2019scar}  & $124.5^{\pm \mathbf{128.6} }$  & $\mathbf{122.9}^{\pm 129.0 }$ & $123.4^{\pm 146.9 }$ & $ 123.4 ^{\pm 197.1 } $ \\
& & SFA-Net \cite{zhu2019dual}  & $\mathbf{128.6}^{\pm \mathbf{133.4} }$ & $128.9^{\pm 162.9 }$ &  $128.7^{\pm 163.2 }$ & $ 128.7 ^{\pm 199.9 } $  \\

\midrule

\multirow{10}{*}[-0.25em]{\rule{0pt}{2ex} Medium }
& \multirow{5}{*}[-0.25em]{\rule{0pt}{2ex} $ \text{\normalsize{STA}}   \atop \mathbf{182} $ } 
& DM-Count \cite{wang2020DMCount} & $\mathbf{88.6}^{\pm \mathbf{64.4 }}$ & $89.6^{\pm 75.9 }$ & $93.0^{\pm 81.3 }$  & $ 64.1^{\pm 78.4} $ \cite{wang2020DMCount}\\
& & BL \cite{ma2019bayesian}  & $\mathbf{68.6}^{\pm 69.9 }$ & $68.9^{\pm 63.3 }$  & $68.7^{\pm \mathbf{61.9} }$  & $63.5 ^{\pm 74.7 }$ \cite{ma2019bayesian}\\
& &  S-DCNet \cite{xhp2019SDCNet} & $66.6^{\pm 72.6 }$ & \cellcolor{SkyBlue}$\mathbf{60.5}^{\pm \mathbf{65.5} }$ & $61.3^{\pm 66.9 }$  & $ 61.3 ^{\pm 88.7 } $   \\
& & SCAR \cite{gao2019scar}  & $83.7^{\pm 67.4 }$ & $\mathbf{72.9}^{\pm \mathbf{61.8} }$ & $79.3^{\pm 67.4 }$ & $ 79.3 ^{\pm 82.9 } $   \\
& & SFA-Net \cite{zhu2019dual}  & $68.4^{\pm 65.1 }$ & $64.9^{\pm 59.5 }$ & $\mathbf{63.6}^{\pm \mathbf{55.6} }$  &  $ 63.6 ^{\pm 92.9 } $ \\

\cmidrule(lr){2-7}

& \multirow{5}{*}[-0.25em]{\rule{0pt}{2ex} $\text{\normalsize{STB}}  \atop \mathbf{316}$} 
& DM-Count \cite{wang2020DMCount} & $9.1^{\pm 9.3 }$ & $\mathbf{8.6}^{\pm \mathbf{8.6} }$ & $8.9^{\pm 10.3 }$  & $ 7.3^{\pm 9.3} $~\cite{wang2020DMCount}\\
& & BL \cite{ma2019bayesian}  & $\mathbf{9.6}^{\pm 9.3 }$ & $9.7^{\pm 9.3 }$ & $10.8^{\pm \mathbf{9.2} }$  & $7.5^{\pm 9.4}$~\cite{ma2019bayesian} \\
& &  S-DCNet \cite{xhp2019SDCNet} &  $9.2^{\pm 9.4 }$ & $9.6^{\pm 10.5 }$ & $\mathbf{7.9}^{\pm \mathbf{8.6} }$  & $ 7.8 ^{\pm 11.0 } $ \\
& & SCAR \cite{gao2019scar}  & $\mathbf{9.8}^{\pm \mathbf{10.2 } }$ & $13.8^{\pm 11.7 }$ & ${10.3}^{\pm 14.0 }$ & $ 10.3 ^{\pm 14.1 } $ \\
& & SFA-Net \cite{zhu2019dual}  & $9.0^{\pm 7.3 }$ & $8.8^{\pm 8.0 }$ & \cellcolor{SkyBlue}$\mathbf{7.4}^{\pm \mathbf{6.8} }$  & $ 7.4 ^{\pm 9.2 } $  \\

\bottomrule
\end{tabular}
}

\end{table*}

\subsection{Ablation Studies}
\label{sec:abl}

For ablation studies, we conducted experiments with DM-Count~\cite{wang2020DMCount} on NWPU dataset. The loss function involved in optimization (Sec.~\ref{sec:stage4}) is of the form 

\begin{equation}
\mathcal{L}^* = \mathcal{L}  + \lambda_2 \widehat{\mathcal{L}} 
\end{equation}

where $\mathcal{L}^*$ is the final loss function, $\mathcal{L}$ is the model loss, $\widehat{\mathcal{L}}$ is the Bin Loss and $\lambda_2$ is a weighting factor. From Eqn.~\ref{eq:8}, we need to tune for both $\lambda_1,\lambda_2$. We conduct a grid optimization with $\lambda_2$ ranging over $\{0.01,1\}$ and $\lambda_1$ over $\{1,10,100\}$. The pooled MAE and standard deviations are summarized in Table~\ref{table:hyp}. Based on the results, we fix $\lambda_1= 1,\lambda_2= 1$ for DM-Count~\cite{wang2020DMCount} on all datasets and minibatching schemes (RR,RS).
\begin{table}[!t]
\captionof{table}{Hyperparameter search for $\lambda_1$ and $\lambda_2$ over a grid and the resulting pooled MAE and standard deviations.}
\label{table:hyp}
\centering
\resizebox{0.75\linewidth}{!}
{
\centering
\begin{tabular}{c|c|c}
\toprule
     $\lambda_1 \downarrow$  $\lambda_2 \rightarrow$ & 0.01 & 1 \\
\toprule
1 & $84.1^{\pm 183.2 }$ & $76.7^{\pm 205.0 }$ \\
10 & $ 80.5^{\pm 243.7 }$ & $79.5^{\pm 238.7 }$ \\
100 & $ 80.7^{\pm 236.8 } $  & $80.4^{\pm 252.7 }$\\
\bottomrule
\end{tabular}
}
\end{table}

The effectiveness of bin-loss (Eqn.~\ref{eq:8}) also depends on the extent to which a reference architecture utilizes the formulation for better optimization. For SCAR~\cite{gao2019scar} and SFA-Net~\cite{zhu2019dual}, we hypothesize that this ability is relatively lower. Therefore, bin-loss is not always better for these networks (see Table~\ref{table:results}). Other networks (BL~\cite{ma2019bayesian}, DM-Count~\cite{wang2020DMCount}) utilize the loss better, leading to consistent improvement in MAE and standard deviation. However, SCAR~\cite{gao2019scar} is still better than no-binning in all cases except NWPU dataset. SFA-Net's performance with bin-loss included is better for the larger UCF, NWPU datasets. Also, inclusion of bin-loss results in consistent gains in terms of error standard deviation especially on the larger, heavily skewed datasets.

As mentioned in Sec.~\ref{sec:partitionprior}, we model the likelihood for each bin as a multinomial distribution. For comparative evaluation, we also consider two other candidate distributions for binning. The first candidate models the likelihood for the bin counts as a Poisson distribution: 

\begin{equation} \label{eqn:poibinlkhood}
\begin{split}
lik(B_k) & = lik(x_1,\dots , x_{m_k} ; \lambda_1, \dots, \lambda_{m_k}) \\
 & = \prod_{j=1}^{m_k} \frac{\lambda_j e^{-\lambda_j}}{x_j}
\end{split}
\end{equation}

where $\lambda_1, \dots, \lambda_{m_k}$ are the parameters of the Poission distributions associated with the bin elements. The other terms are used in the same context as Eqn.~\ref{eqn:binlkhood} in Section~\ref{sec:partitionlkhood}. The second candidate distribution for binning is a variant of the multinomial, called stratified multinomial~\cite{sbb}. In this variant, the optimal Bayesian binning is applied not only to the count range, but also to the count frequency distribution. The comparative results can be seen in Table~\ref{table:likelihoodablation}. Though the pooled MAE with Poisson binning is slightly lower for random binning, the standard deviation is significantly larger than in the case of multinomial (as employed by us). The other results indicate the better overall stability arising from our simple yet effective choice for the likelihood distribution. 

\begin{table}[!t]
\captionof{table}{Ablations on the likelihood model for different choices of bin-level distribution. Though the pooled MAE with Poisson distribution is slightly lower for random binning, the standard deviation is significantly larger than our choice (multinomial).}
\label{table:likelihoodablation}
\centering
\resizebox{\linewidth}{!}
{
\centering

\begin{tabular}{c|c|c|c}
\toprule
     $Likelihood\downarrow$  Binning$\rightarrow$ & Bin Loss & Bin Loss (RR) & No-binning\\
\toprule
Poisson & $84.8^{\pm 441.2 }$ & $89.1^{\pm 533.1 }$ &  $77.8^{\pm 380.3 }$ \\
Stratified Multinomial & $90.0^{\pm 283.5 }$  & $90.6^{\pm 374.0 }$ & $80.7^{\pm 290.7 }$  \\
\textbf{Multinomial (ours)} &  $88.1^{\pm 236.7 } $ & $76.7^{\pm 205.0} $ & $77.8^{\pm 214.9 }$\\
\bottomrule
\end{tabular}
}
\end{table}

\section{Conclusion}

In this paper, we highlight biases at various stages of the typical crowd counting pipeline and propose novel modifications to address issues at each stage. We propose a novel Bayesian sample stratification approach to enable balanced minibatch sampling. Complementary to our sampling approach, we propose a novel loss function to encourage strata-aware optimization. We analyze the performance of crowd counting approaches across standard datasets and demonstrate that our proposed modifications reduce error standard deviation in a noticeable manner. Altogether, our contributions represent a nuanced, statistically balanced and fine-grained characterization of performance for crowd counting approaches.

The proposed bin-aware loss visibly reduces standard deviation of error. However, our work highlights the need for approaches in which error deviations are negligible compared to the mean error. We hope that our work motivates the community to join us in exploring these challenging aspects of the problem. Studying and addressing issues we have raised would enable statistically reliable crowd counting approaches in future.

\bibliographystyle{ACM-Reference-Format}
\bibliography{egbib}

\end{document}